\documentclass{article}

\usepackage{microtype}
\usepackage{graphicx}
\usepackage{subcaption}
\usepackage{booktabs} %

\usepackage{xurl}
\usepackage{hyperref}

\usepackage[preprint]{icml2026}

\usepackage{amsmath}
\usepackage{amssymb}
\usepackage{mathtools}
\usepackage{amsthm}
\usepackage{enumitem}
\setlist[itemize]{noitemsep, nolistsep}
\usepackage[raggedrightboxes]{ragged2e}
\usepackage{makecell}
\usepackage{tabularx}
\newcolumntype{b}{X}
\newcolumntype{s}{>{\hsize=.5\hsize}X}

\usepackage[capitalize,noabbrev]{cleveref}

\theoremstyle{plain}

\theoremstyle{definition}

\theoremstyle{remark}

\usepackage[textsize=tiny]{todonotes}
\usepackage{bbm}
\usepackage{fancyvrb} %
\VerbatimFootnotes

\icmltitlerunning{Omitted Variable Bias in Language Models Under Distribution Shift}

\begin{document}

\twocolumn[
  \icmltitle{Omitted Variable Bias in Language Models Under Distribution Shift}

  \icmlsetsymbol{equal}{*}

  \begin{icmlauthorlist}
    \icmlauthor{Victoria Lin}{cmu}
    \icmlauthor{Louis-Philippe Morency}{cmu}
    \icmlauthor{Eli Ben-Michael}{cmu}
  \end{icmlauthorlist}

  \icmlaffiliation{cmu}{Carnegie Mellon University, Pittsburgh, PA, USA}

  \icmlcorrespondingauthor{Victoria Lin}{vlin2@andrew.cmu.edu}

  \icmlkeywords{omitted variable bias, language models, distribution shift}

  \vskip 0.3in
]

\printAffiliationsAndNotice{}  %

\begin{abstract}
Despite their impressive performance on a wide variety of tasks, modern language models remain susceptible to distribution shifts, exhibiting brittle behavior when evaluated on data that differs in distribution from their training data. In this paper, we describe how distribution shifts in language models can be separated into \textit{observable} and \textit{unobservable} components, and we discuss how established approaches for dealing with distribution shift address only the former. Importantly, we identify that the resulting \textit{omitted variable bias} from unobserved variables can compromise both evaluation and optimization in language models.
To address this challenge,
we introduce a framework that maps the strength of the omitted variables to bounds on the \textit{worst-case generalization performance} of language models under distribution shift. In empirical experiments, we show that using these bounds directly in language model evaluation and optimization provides more principled measures of out-of-distribution performance, improves true out-of-distribution performance relative to standard distribution shift adjustment methods, and further enables inference about the strength of the omitted variables when target distribution labels are available.

\end{abstract}

\section{Introduction}

For language models to be widely useful in real-world applications, they must remain robust, reliable, and performant across data distributions and contexts. However, even for modern large language models (LLMs) capable of impressive performance on diverse tasks, the challenge of \textit{distribution shift} remains. 
When evaluated on data that differs in distribution from their training data, LLMs often exhibit brittle behavior, such as difficulty answering questions containing simple mutations that do not appear in training \cite{xu2025reimagine, huang2025mathperturb} or failures in reasoning given counterfactual information that alters the question's world state \cite{NEURIPS2024_d5a1f97d, huyuk2025reasoning}.

Established approaches for dealing with distribution shift in language models account for shifts that are \textit{observable} to language models---that is, shifts resulting from variables that models are able to capture from the text. We place a novel focus on a component of distribution shift that these existing methods fail to address: the \textit{unobservable} component of distribution shift, or shifts over variables that cannot be captured directly by language models.  This non-observability may occur for two reasons. (i) The shift may occur over variables that are external to the text, such as attributes of the individuals who wrote or labeled the texts. 
These variables typically remain unmeasured in observational text settings.
(ii) The shift may occur over information that is contained in the \textit{text} but not in a model's \textit{representation} of the text.
While this information is observable to a human reader, 
it is not observable to a language model. This type of information loss is significant, since it almost inevitably occurs in language data as nearly infinite-dimensional, unstructured raw text is reduced to a lower-dimensional numerical form that can be used by models.

In this paper, we first identify that the \textit{omitted variable bias} (OVB)  arising from these unobserved variables can compromise the evaluation and optimization of language models under distribution shift. In evaluation, failure to account for omitted variables can lead to overly optimistic performance estimates in the target domain; in optimization, it can yield models that still remain brittle despite adjustment for observable sources of distribution shift. 

To address these challenges, we introduce a framework that maps the strength of the omitted variables to a bound on a language model's generalization performance. Under this framework, although the omitted variables themselves are unobserved, their potential influence can be benchmarked 
and
used to identify a set of plausible target distributions over which the \textit{worst-case generalization performance} of a language model can be defined.

This worst-case generalization bound gives rise to three primary contributions, which we demonstrate empirically. First, the bound provides a more principled and robust metric for evaluating generalization performance in the absence of target distribution labels, improving on standard adjustment objectives that account only for observed distribution shift. Second, the bound can be directly optimized to produce models that are explicitly more robust to unobserved sources of shift and that generalize more reliably to the target distribution.
Third, when target labels \textit{are} available and true test performance can be computed, the bound can be used to infer the strength of the omitted variables for a given model under distribution shift, offering interpretability to models that are otherwise opaque.

\section{Related Work}

\paragraph{Domain adaptation and distribution shift.} Model performance degradation resulting from differences between source (training) and target (test) distributions has long been studied in the machine learning literature. These distribution shifts arise from a variety of factors, such as differing domains or disparities among subgroups \citep{arjovsky2020invariantriskminimization, 10.5555/3618408.3620060}. Even for modern LLMs, selection biases when collecting data for preference alignment may skew the distribution of human feedback during post-training \citep{lin2024optimizing}; and recent work on reasoning suggests that LLMs' strong performance may be at least partially attributed to memorization of their training data, as even minor perturbations in test distributions can elicit large drops in performance \citep{xu2025reimagine, huang2025mathperturb}.

One of the most common and portable methods for addressing distribution shift is \textit{importance weighting}, which reweights training samples to match the target distribution via a density ratio. Importance-weighted estimators are often combined with outcome models to form \textit{doubly robust} estimators that are more robust to misspecification and estimation error \citep{Robins1994}. These methods are widely used to correct for covariate and label shift across prediction, recommendation, and reinforcement learning tasks \citep{pmlr-v97-byrd19a, pmlr-v162-kallus22a, NEURIPS2022_e124f154, lin2024optimizing}. However, when relevant variables are omitted, bias can arise in both the density ratio and outcome model, threatening the validity of these adjustments.

\paragraph{Omitted variable bias.} In practice, OVB can be difficult to estimate, as computing it directly requires knowing which variables have been omitted from the unknown full set of variables.
Moreover, a particular challenge for language models is that the ``full set of variables'' corresponds to the raw text used as input---but any method modeling text still requires that it be represented numerically, and so the estimator loses access to the omitted variables. 

Due to the difficulties of computing OVB directly, a long history of work has set out to provide bounds on the degree of OVB under various (generally linear) modeling assumptions \citep{goldberger1991course, frank2000impact, angrist2009mostly, oster2019unobservable, cinelli2019making}. To address the limitations of these strong assumptions, recent work from \citet{chernozhukov2024longstoryshortomitted} establishes bounds on the OVB of causal parameters even when using nonparametric models. Subsequent work from \citet{lin2025isolated} 
extends these bounds to a language setting where the goal is to estimate causal effects from language. 
However, because most language tasks are not explicitly causal, the question of how to 
extend bounds and other theoretical properties of omitted variables to language models becomes less clear.
Our work addresses this challenge by mapping the strength of omitted variables to bounds on the generalization performance of 
language models under a general distribution shift setting.

\section{Proposed Approach}

\label{sec:approach}

\subsection{A General Form for Doubly Robust Losses}
\label{sec:dr_general}

We consider a distribution shift setting where the source distribution $P$, from which training data is derived, may differ from the target distribution $Q$. Then the \textit{generalization performance} of a model $f$ under this setting is given by its performance in the target distribution, or
\[
\mathcal{L} = \mathbb{E}_Q[\ell(Y, X; f)]
\]
where $X$ are texts, $Y$ are labels, and $\ell$ is the model loss.

To learn a model with good generalization performance, the goal is to minimize this quantity. In a distribution shift setting, however, labels are available only for the source distribution $P$ and not the target distribution $Q$. That is, we observe a labeled dataset from $P$:
\begin{equation*}
    D_P=\{(X_i, Y_i)\sim P\}, \; i \in [n] \quad X_i \in \mathcal{X}, Y_i \in \mathbb{R}
\end{equation*}

and an unlabeled dataset from $Q$.
\begin{equation*}
    D_Q=\{X_j\sim Q\}, \; j \in [m] \quad X_j \in \mathcal{X}
\end{equation*}

\paragraph{Identification.} Since $\mathcal{L}$ is written in terms of the unobserved labels $Y \sim Q$, it cannot be directly computed.
We will consider a setting where it is possible in principle to identify $\mathcal{L}$ in terms of the observed distribution $P$ given access to the text $X$ and potentially additional covariates $Z$ that are external to the text and that we do not observe.

The following assumptions are standard assumptions for domain shift, if we had access to these additional external values $Z$. (i) \textbf{Covariate shift.} We assume a covariate shift setting where---given access to the full set of variables---the relation $\mathbb{E}_P[\ell(Y, X; f) \mid X, Z] = \mathbb{E}_Q[\ell(Y, X; f) \mid X, Z]$ holds. (ii) \textbf{Overlap.} We overlap between the source and target distributions: $\frac{dQ}{dP}(X, Z) = \frac{\text{Pr}_Q(X,Z)}{\text{Pr}_P(X,Z)}$ is finite for all pairs of $X$ and $Z$. In intuitive terms, this assumption ensures that any text with a non-zero probability of appearing in $Q$ has also a non-zero probability of appearing in $P$.

If we had access to the unobserved external variables $Z$, then letting 
$\alpha(X, Z)=\frac{\text{Pr}_Q(X,Z)}{\text{Pr}_P(X,Z)}$
be the Riesz representer,
a common approach for distribution shift adjustment is to compute the \textit{importance-weighted} objective:
\[
\mathcal{L}_{\text{IPW}} = \mathbb{E}_P\left[\alpha(X, Z)\ell(Y, X; f)\right]
\]

The addition of an outcome model 
$g(h(X), X, Z; f) = \mathbb{E}[\ell(Y, X; f) \mid X, Z)]$
then yields the following \textit{doubly robust} objective. Note that here $g$ is a predictor of the loss $\ell$ itself rather than the label $Y$.
\begin{align*}
\mathcal{L}_{\text{DR}} &= \mathbb{E}_P\left[\alpha(X,Z)(\ell(Y, X; f) - g(X, Z;f))\right] \\
&\quad +\mathbb{E}_Q[g(X, Z; f)]
\end{align*}
Both $\mathcal{L}_{\text{IPW}}$ and $\mathcal{L}_{\text{DR}}$ are mathematically equivalent to the original $\mathcal{L}$ but allow the loss to be computed over the observed data $D_P$. $\mathcal{L}_{\text{DR}}$ further confers robustness against misspecification of the density ratio $\frac{\text{Pr}_Q}{\text{Pr}_P}$ as long as the outcome model $g$ is correct, or vice versa. Derivations for these objectives are found in Appendix \ref{sec:dr_derivation}.

\paragraph{Estimation.} In practice, any estimator of the doubly robust objective will have access only to variables observable to language models. Notably, in the language model setting, these observable variables correspond only to $h(X)$, the language representation of the text $X$, where $h(\cdot): \mathcal{X} \rightarrow \mathbb{R}^d$, and do not include either the full text or the unobserved external variables $Z$.

Following the terminology used in \citet{chernozhukov2024longstoryshortomitted} to refer to models learned over partial feature sets, we first define the ``short'' outcome model $g(h(X); f) = \mathbb{E}_P[\ell(Y, X; f) \mid h(X)]$  and ``short'' Riesz representer $\alpha(h(X))= \frac{\text{Pr}_Q(h(X))}{\text{Pr}_P(h(X))}$, which are learned given only the text representation $h(X)$. These are contrasted with the ``long'' outcome model and ``long'' Riesz representer, which are learned over the 
full text $X$ and unknown external variables $Z$ (and therefore also implicitly over the representation $h(X)$, which is a deterministic function of $X$). 

Together, these yield an estimator of the short doubly robust objective $\mathcal{L}_{\text{DR}_s}$ that replaces the long models with the short models:
\begin{align*}
\widehat{\mathcal{L}}_{\text{DR}_s} &= \widehat{\mathbb{E}}_P\left[\widehat{\alpha}(h(X))(\ell(Y, X; f) - \widehat{g}(h(X) ;f)\right] \\
&\quad+\widehat{\mathbb{E}}_Q[\widehat{g}(h(X); f)]
\end{align*}
where $\widehat{g}$ is estimated on a held-out sample from $P$, and $\widehat{\alpha}$ is estimated on a held-out sample of texts from both $P$ and $Q$. $\widehat{\mathbb{E}}_P$ represents the sample average in the dataset drawn from the source distribution $P$.

Adjustment using only the model-observable variables $h(X)$ accounts for only the first component of distribution shift.
In essence, it assumes that no information in the text $X$ other than the representation $h(X)$ is systematically different between the source and the target. It also assumes that the unobserved external variables $Z$ are not systematically different.
Concretely, this assumes that the covariate shift assumption holds when conditioning only on $h(X)$, rather than the full set of variables $X$ and $Z$.

When this assumption fails and there are systemic differences between the source and target distributions not captured solely by the representation $h(X)$, OVB is introduced in the short doubly robust objective $\mathcal{L}_{\text{DR}_s}$, and the estimator $\widehat{\mathcal{L}}_{\text{DR}_s}$ may no longer accurately reflect generalization performance in the target distribution $Q$. As a result, models learned using this objective may insufficiently account for distribution shift and perform poorly when deployed outside their training domain.

\subsection{Worst-Case Generalization Performance}

Without access to the full covariate set $(X,Z)$, unbiased estimation of $\mathcal{L}_{\text{DR}}$ is no longer possible under standard identification assumptions. Relying only on the representation $h(X)$ yields a partially identified generalization objective whose value depends on unobserved variables omitted from the distribution shift adjustment set.  We propose to account for this uncertainty by incorporating the omitted variable bias induced by the unobserved covariates \citep{chernozhukov2024longstoryshortomitted}.

Without full knowledge of the covariates, the observed data define a set of plausible target distributions over which the model's performance may vary. To account for this uncertainty, we adopt a distributionally robust optimization (DRO) framework \citep{duchi2021dro} that defines the doubly robust objective under the worst-case distribution with respect to OVB within this set of plausible targets. This approach yields a \textit{bound on the model's generalization performance} that provides robustness guarantees under unobserved distribution shift. We therefore expect that models evaluated using this worst-case objective will more accurately reflect performance in the target distribution, and models learned using this objective will generalize more reliably to the target distribution than those trained using the standard doubly robust objective.

\paragraph{OVB definition.} Extending \citet{chernozhukov2024longstoryshortomitted}, the OVB of $\mathcal{L}_{\text{DR}_s}$ is bounded as:
\begin{equation*}
    |\underbrace{\mathcal{L}_{\text{DR}_s}-\mathcal{L}_{\text{DR}}}_{\text{OVB}}|^2 \leq \rho^2 C_Y^2C_D^2\sigma^2\nu^2 ,
\end{equation*}

where the \textit{fidelity} $\sigma^2$ and the \textit{overlap} $\nu^2$ are identifiable directly from the data as
\begin{align*}
    \sigma(f)^2 & =\mathbb{E}_P[(\ell(Y,X;f)-g(h(X);f))^2]\\
    \nu^2& =\mathbb{E}_P[\alpha(h(X))^2].
\end{align*}
The fidelity measure $\sigma^2$ indicates how well the short outcome model predicts the loss, while the overlap measure $\nu^2$ indicates how well the overlap assumption between $P$ and $Q$ is fulfilled by the short Riesz representer.

The OVB bound furthers depend on the parameters $C_Y$, $C_D$, and $\rho$, where $C_Y$ and $C_D$ denote the explanatory power of the omitted variables toward the outcome model and Riesz representer, respectively, and $\rho$ is the degree of confounding present. These are defined as
\[
C_Y(f)^2=\frac{\mathbb{E}_P[(g(X, Z;f) - g(h(X);f))^2]}{\mathbb{E}_P[(\ell(Y,X;f) - g(h(X);f))^2]}
\]
\[
C_D^2=\frac{\mathbb{E}_P[\alpha(X, Z)^2]-\mathbb{E}_P[\alpha(h(X))^2]}{\mathbb{E}_P[\alpha(h(X))^2]}
\]
\begin{align*}
    \rho(f)=\text{Corr}^2(&g(X, Z;f) - g(h(X);f), \\
    &\alpha(X, Z) - \alpha(h(X)))
\end{align*}

Because the long outcome model $g(X, Z;f)$ and the long Riesz representer $\alpha(X, Z)$ are not known (except in an oracle setting), these quantities are not identified directly from the data. Instead, they serve as sensitivity parameters to characterize the level of worst-case OVB.

\paragraph{Worst-case objective.} Under the OVB bounds, the observed data admit a family of target distributions characterized by $C_Y$,  $C_D$, and $\rho$. Letting $f \in \mathcal{F}$ denote a model from a fixed hypothesis class, we define:
\[
C_Y^{\max} := \sup_{f \in \mathcal F} C_Y(f)
\quad
\rho^{\max} := \sup_{f \in \mathcal F} \rho(f)
\quad
C_D^{\max} := C_D
\]

Then we define the set of possible target distributions as
\[
\mathcal Q
=
\left\{
\tilde Q :
C_Y \le C_Y^{\max},
C_D \le C_D^{\max},
\rho \le \rho^{\max}
\right\}
\]

As we mention previously, $C_Y$, $C_D$, and $\rho$ (and consequently $C_Y^{\max}$, $C_D^{\max}$, and $\rho^{\max}$) cannot be identified directly from the data---but they \textit{can} be benchmarked to observed data. The resulting benchmarked values can then be used to define a plausible range for these sensitivity parameters, as we later show in our empirical experiments.

Then for a fixed model $f \in \mathcal{F}$, the worst-case generalization performance\footnote{Confidence intervals on the OVB can also be computed following \citet{chernozhukov2024longstoryshortomitted}. These can be used directly to define a confidence bound on worst-case generalization performance.} over this set is given by
\[
\sup_{\tilde Q \in \mathcal Q}
\mathbb{E}_{\tilde Q}\!\left[\ell(Y,X;f)\right]
=
\mathcal L_{\text{DR}}
+
\rho^{\max} C_Y^{\max}C_D^{\max}\sigma\nu
\]

Evaluating models with this performance metric directly leads to our first contribution: the ability to robustly evaluate the generalization performance of language models in the target domain.

Next, by minimizing this quantity, i.e., solving
\[
\min_{f\in\mathcal{F}} \sup_{\tilde{Q} \in \mathcal{Q}}  \mathbb{E}_{\tilde{Q}}[\ell(Y, X; f)]
\]
we arrive at our second contribution: a model optimized in this way should generalize more reliably to the target distribution because it explicitly accounts for the uncertainty introduced by unobserved sources of distribution shift.

Finally, the sensitivity parameters themselves bring us to our third contribution. If a model's performance on the target distribution can be computed directly---for example, if we have access to some target labels---then it is possible to infer an empirical upper bound on $\rho$, $C_Y$, and $C_D$ by tuning the sensitivity parameters until the worst-case generalization performance matches the observed test performance. Comparing these sensitivity parameters among language models provides useful insight into the strength of the omitted variables for each model: larger values of the sensitivty parameters correspond to more explanatory power coming from the omitted variables. This in turn signals whether or not a model has captured the concepts that distinguish the source and target distribution, conferring a degree of interpretability on models that otherwise might have none.

\begin{table*}[!ht]
    \centering
    \caption{NLP tasks with GLM log-likelihood losses and their correspondence to language model outputs $f(x)$.}
    \resizebox{1.0\linewidth}{!}{
    \begin{tabular}{|c|c|c|c|c|}
    \toprule
    & $\eta$ & $b(\eta)$ & Label & Mean parameter \\
    \midrule
    \textbf{Regression} & $f(x) \in \mathbb{R}$ & $\frac{1}{2}\eta^2$ & $Y \in \mathbb{R} $& $\mu=\eta$ \\
    \textbf{Binary classification} & $f(x) \in \mathbb{R}$ (logit) & $\log(1+\exp(\eta))$ & $Y \in \{0, 1\}$ & $p=\sigma(\eta)$ \\
    \textbf{Multiclass classification} & \makecell{$f(x) \in \mathbb{R}^K$ \\ (logits)} & $\log \sum_{j=1}^K \exp(\eta_j)$ & $Y \in \{0,1\}^K$ & $p_j=\frac{\exp(\eta_j)}{\sum_{k \in K}\exp(\eta_k)}$ \\
    \textbf{Text generation} & \makecell{$f^{(t)}(x) \in \mathbb{R}^K$ \\ (logits at step $t$)} & $\sum_{t=1}^T \log\sum_{j=1}^K \exp(\eta_j^{(t)})$ & $Y^{(t)} \in \{0,1\}^K$ & $p_j^{(t)}=\frac{\exp(\eta_j^{(t)})}{\sum_{k \in K}\exp(\eta_k^{(t)})}$ \\
    \bottomrule
    \end{tabular}
    }
    \label{tab:glm_losses}
\end{table*}

\section{GLM Log-Likelihood Losses}

Our formulation can be used to bound worst-case generalization performance for any generic task loss. However, it
requires that the model $g$ predict the loss itself as the conditional outcome rather than the label. This may make optimization more challenging for complex losses, as the loss includes the model $f$ that is being optimized. 

In this section, we consider the class of generalized linear model (GLM) log-likelihood losses, which include tasks considered to be core to natural language processing such as regression, classification, and text generation. Due to their log-linear structure, these losses allow for an alternative implementation of the bound on generalization performance where the outcome model simply predicts the label. This results in a much easier optimization problem that is widely useful for many common tasks.
Derivations and technical results for this section are deferred to Appendix \ref{sec:glm_derivations} and \ref{sec:glm_losses}.

\subsection{A Worst-Case Objective for GLM Losses}

Following the generic form of a GLM negative log-likelihood, let
\[
\ell(Y, X; \eta, f)=-(Y\cdot \eta(X; f) - b(\eta(X; f)) + c(Y))\
\]

where $\eta(\cdot)$ is the natural parameter, $b(\cdot)$ is the log-partition function, and $c(Y)$ is a constant with respect to optimization and therefore can be ignored. Then we have the losses:
\[
    \mathcal{L}=\mathbb{E}_Q[-Y\cdot \eta(X; f) + b(\eta(X;f))]
\]
\[
    \mathcal{L}_{\text{IPW}}=\mathbb{E}_Q[b(\eta(X;f))] - \mathbb{E}_P\left[\frac{\text{Pr}_Q(X,Z)}{\text{Pr}_P(X,Z)}\eta(X;f)Y\right] 
\]

This yields an alternative decomposition of the doubly robust objective in which the portion of the loss that is dependent on $f$ is contained in the Riesz representer $\alpha$ rather than the outcome model $g$:
\[
\alpha(X, Z; f)=\frac{\text{Pr}_Q(X,Z)}{\text{Pr}_P(X,Z)}\eta(X; f)
\]

The GLM log-likelihood loss allows us to define the conditional outcome as simply the label $Y$, and we can relax our covariate shift assumption compared to the requirement for the general form in Section \ref{sec:dr_general}. Here, we assume simply that $\mathbb{E}_P[Y \mid X, Z]=\mathbb{E}_Q[Y \mid X, Z]$. Letting $g$ now be a predictor of the label $Y$ rather than the loss, i.e., $g(X, Z)=\mathbb{E}[Y \mid  X, Z]$, we have
\begin{align*}
    \mathcal{L}_{\text{DR}}&=\mathbb{E}_Q[b(\eta(X;f))-\eta(X;f)g(X, Z)] \\
    &\quad-\mathbb{E}_P\left[\alpha(X, Z; f)(Y-g(X, Z))\right]
\end{align*}

and as before, the worst-case generalization performance of model $f$ over the plausible target distribution set $\mathcal{Q}$ is
\[
\sup_{\tilde Q \in \mathcal Q}
\mathbb{E}_{\tilde Q}\!\left[\ell(Y,X;f)\right]
=
\mathcal L_{\text{DR}}
+
\rho^{\max} C_Y^{\max}C_D^{\max}\sigma\nu
\]

\paragraph{Task-specific parameters.} From this general form, the loss functions and worst-case task objectives for regression, classification, and text generation are easily obtained by using the correct natural parameter $\eta$ and log-partition function $b$ for each task (we identify these in Table \ref{tab:glm_losses}), then substituting into the GLM worst-case objective.

\subsection{From Long to Short}

Under the alternative decomposition of the doubly robust objective, the short outcome model and Riesz representer are now given by $g(h(X))=\mathbb{E}_P[Y \mid h(X)]$ and $\alpha(h(X); f)=\frac{\text{Pr}_Q(h(X))}{\text{Pr}_P(h(X))}\eta(h(X);f)$, respectively. From these, the fidelity $\sigma^2$ and overlap $\nu^2$ follow:
\[
\sigma^2=\mathbb{E}_P[(Y-g(h(X)))^2] \qquad \nu^2=\mathbb{E}_P[\alpha(h(X);f)]
\]

and all other
parameters associated with the worst-case objective are defined as before.

\section{Experiments}

In our experiments, we study three use cases of the worst-case generalization bound aligned with our contributions. First, we use the bound for evaluation and show that ignoring unobserved variables leads to overly optimistic estimates of target performance. Second, we demonstrate how the bound can quantify the strength of omitted variables even for a black-box commercial LLM on a complex reasoning task. Finally, across multiple datasets and tasks, we show that optimizing models for worst-case generalization improves out-of-distribution performance relative to standard doubly robust methods that assume no missing information.

\subsection{Datasets}

\paragraph{Math Reasoning (text generation, real-world setting).} Our Math Reasoning dataset pairs the original MATH dataset \citep{hendrycks2021measuring} and a recent variation, MATH-Perturb \citep{huang2025mathperturb}, both consisting of problems from high school math competitions. 
MATH-Perturb perturbs existing problems in the MATH dataset, with two versions---MATH-P-Hard and MATH-P-Simple---that do or do not respectively change the underlying logic of the original question. We treat MATH as the source distribution and each variant of MATH-Perturb as a target distribution. This dataset is particularly appropriate for evaluating OVB in LLMs, since MATH-Perturb was developed in direct response to concerns that LLMs' good benchmark performance on MATH is due to memorization of the dataset in training, and its authors show that LLMs perform significantly worse on MATH-Perturb than on MATH.

\paragraph{Amazon (regression, semi-synthetic setting).} The Amazon dataset \citep{mcauley2013amazon} consists of customer product reviews, each associated with ``helpfulness'' votes from other customers. We take reviews of musical instruments and office products as our respective source and target distributions. To ensure control of the omitted variables, we encode the reviews with the LIWC lexicon \citep{pennebaker2015development} and generate a new continuous semi-synthetic label $Y$ using the 10 LIWC features most predictive of the helpful vote count. From this, variables can be intentionally omitted by restricting a model's access to only a subset of these 10 features at training time.

\paragraph{EmoBank (regression, real-world setting).} The EmoBank dataset \citep{buechel2017emobank} consists of sentences labeled by human annotators according to their valence, or the positivity or negativity of the text, on a 5-point scale ($Y$). We create two splits of this dataset over their \textit{writer-intended} valence, taking texts with high and low writer-intended valence as the source and target distributions, respectively. Importantly, writer-intended valence differs from the label, which is the annotator's \textit{perceived} valence, but is correlated with it. This creates a natural data setting where an obvious variable likely to induce differences between the source and target distributions---the writer's intent---is not fully observable. 

\paragraph{Hate Speech (classification, real-world setting).} The Hate Speech dataset \citep{qian-etal-2019-benchmark} consists of user comments from the social media sites Reddit and Gab. Each comment is labeled by a human annotator with a binary indicator of hate speech ($Y$). 
We use Reddit as the source distribution and Gab as the target distribution. This dataset again allows us to evaluate a natural setting where OVB may arise, as some variables explaining differences between Reddit and Gab comments may not be observable (e.g., factors that impact both a user's choice of site to use and the type of content the user tends to post).

\subsection{Language Model Evaluation and Measuring Omitted Variables}

In our first two sets of experiments, we use the worst-case generalization bound to evaluate model performance and measure the strength of omitted variables on the complex Math Reasoning dataset. We consider a setting where we have access to labels in the target distribution.

We take pre-trained GPT-4.1 \citep{openai2024gpt4technicalreport} as our language model $f$, with an expectation based on \citet{huang2025mathperturb} that it has already seen the source distribution MATH in its training data but not the target distribution MATH-Perturb. To fit the nuisance parameters $\widehat{g}$ and $\widehat{\alpha}$, we use simple models based on the LIWC and Empath \citep{10.1145/2858036.2858535} lexicons; earlier-generation transformers BERT \citep{devlin-etal-2019-bert}, RoBERTa \citep{liu2019robertarobustlyoptimizedbert}, and MiniLM \citep{NEURIPS2020_3f5ee243}; and GPT-4.1 itself.

For each model-specific generalization bound, we estimate the corresponding upper bound on the OVB sensitivity parameters, $\rho C_YC_D$, by tuning them until the bound intersects the true test performance. We also compute $\rho C_YC_D$ at the intersection between the confidence interval of worst-case generalization performance and the true test performance. Together, these intersections define a reasonable range for the sensitivity parameters, which characterize the strength of the omitted variables. Comparing these values across models therefore indicates the degree to which each model is susceptible to OVB from unobservable distribution shifts.

\subsection{Language Model Optimization}

\paragraph{Semi-synthetic setting.} We recall that in the semi-synthetic Amazon setting, we generate the label $Y$ from 10 known LIWC features and omit variables by providing incomplete subsets of those 10 features to models at train time. Because both the ``long'' and ``short'' feature sets are known, all values required to compute the worst-case generalization bound---including the true sensitivity parameters---can be identified directly from the data. For 100 iterations, we omit a random number of random features and train a model $f$ using the worst-case generalization bound, then compute the test performance of the model on the target distribution.

\paragraph{Real-world settings.} For the EmoBank dataset, comments are encoded using LIWC, and for the Hate Speech dataset, comments are encoded using the SenteCon lexicon, which provides more contextual information in its representations compared to LIWC \citep{lin-morency-2023-sentecon}. 

In these real-world settings, the true sensitivity parameters are not known and must therefore be chosen by the user. We optimize models for the worst-case generalization bound while iterating over a set of plausible parameter values, where larger values correspond to a more pessimistic bound. Building on precedents from the sensitivity analysis literature \citep{cinelli2019making, oster2019unobservable}, we benchmark these values by computing estimates for the sensitivity parameters using a proxy ``long'' model trained on comments encoded using higher-dimensional representations from the transformer MPNet \citep{NEURIPS2020_c3a690be}. When such higher-dimensional representations are unavailable (e.g., for LLMs), analogous benchmarks can be obtained by treating the existing representation as the long model and intentionally omitting features to construct a proxy “short” model, from which sensitivity parameters can be estimated.

\paragraph{Implementation.} We compare our approach to two baselines: a model optimized using the doubly robust objective $\widehat{\mathcal{L}}_{DR}$ (\textbf{DR}), and a model optimized without any distribution shift adjustment (\textbf{unadjusted}). We train our nuisance parameters $\widehat{g}$ and $\widehat{\alpha}$ on held-out samples consisting of 30\% of the training data. Due to the low-dimensional feature set, we use small MLPs with a single hidden layer of size 100 for both nuisance models, which we implement in \texttt{scikit-learn} using default hyperparameters. We use a linear model for the main model $f$ and optimize with \verb|scipy|.

\section{Results and Discussion}

\subsection{Language Model Evaluation}

\begin{figure}[!t]
    \centering
    \begin{subfigure}[t]{0.51\columnwidth}
        \centering
        \includegraphics[width=\linewidth,   trim=10 0 10 0,
  clip]{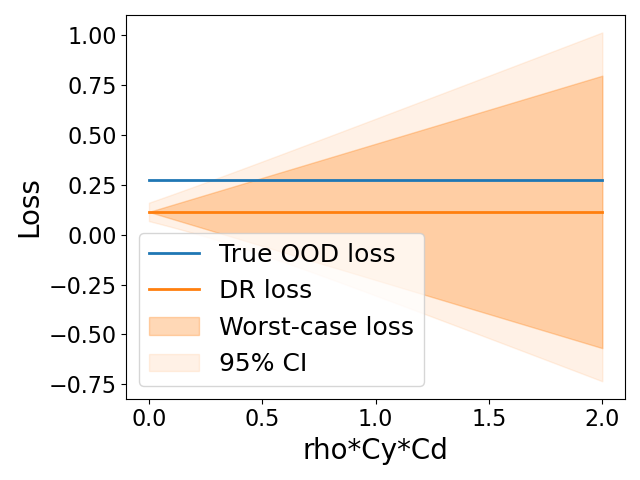}
        \caption{MATH-P-Simple}
        \label{fig:math_simple_eval}
    \end{subfigure}
    \hfill
    \begin{subfigure}[t]{0.464\columnwidth}
        \centering
        \includegraphics[width=\linewidth,   trim=50 0 10 0,
  clip]{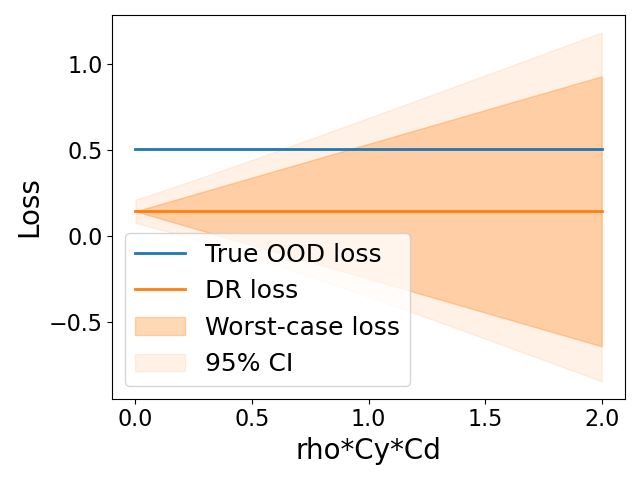}
        \caption{MATH-P-Hard}
        \label{fig:math_hard_eval}
    \end{subfigure}
    \caption{Evaluations of true and distribution shift-adjusted performance on MATH-Perturb.}
    \label{fig:math_eval}
\end{figure}

In Figures \ref{fig:math_simple_eval} and \ref{fig:math_hard_eval}, we plot both the true performance in the target domain and the estimated performance using a standard doubly robust objective, which adjusts only for observed distribution shift. We see that for both MATH-P-Simple and MATH-P-Hard, the doubly robust objective does not recover true performance---though the gap is smaller for the former. This is consistent with expectations, as the problems in MATH-P-Hard contain logic that differs systematically from that of MATH, while MATH-P-Simple varies only on surface-level attributes like the numbers provided in the problem. Importantly, in both cases the doubly robust objective is \textit{optimistic} relative to the true test performance.

Looking to our generalization bound, however, we see 
that as we increase the assumed strength of the omitted variables, the worst-case performance intersects with the true test performance---indicating that if plausible values are chosen for the sensitivity parameters (e.g., by benchmarking as we previously describe), then accounting for unobserved variables via our bound will provide a more accurate picture of performance in the target distribution.

\subsection{Strength of Omitted Variables}

\begin{figure}[!t]
    \centering
    \includegraphics[width=0.9\columnwidth, trim=0 770 0 0, clip]{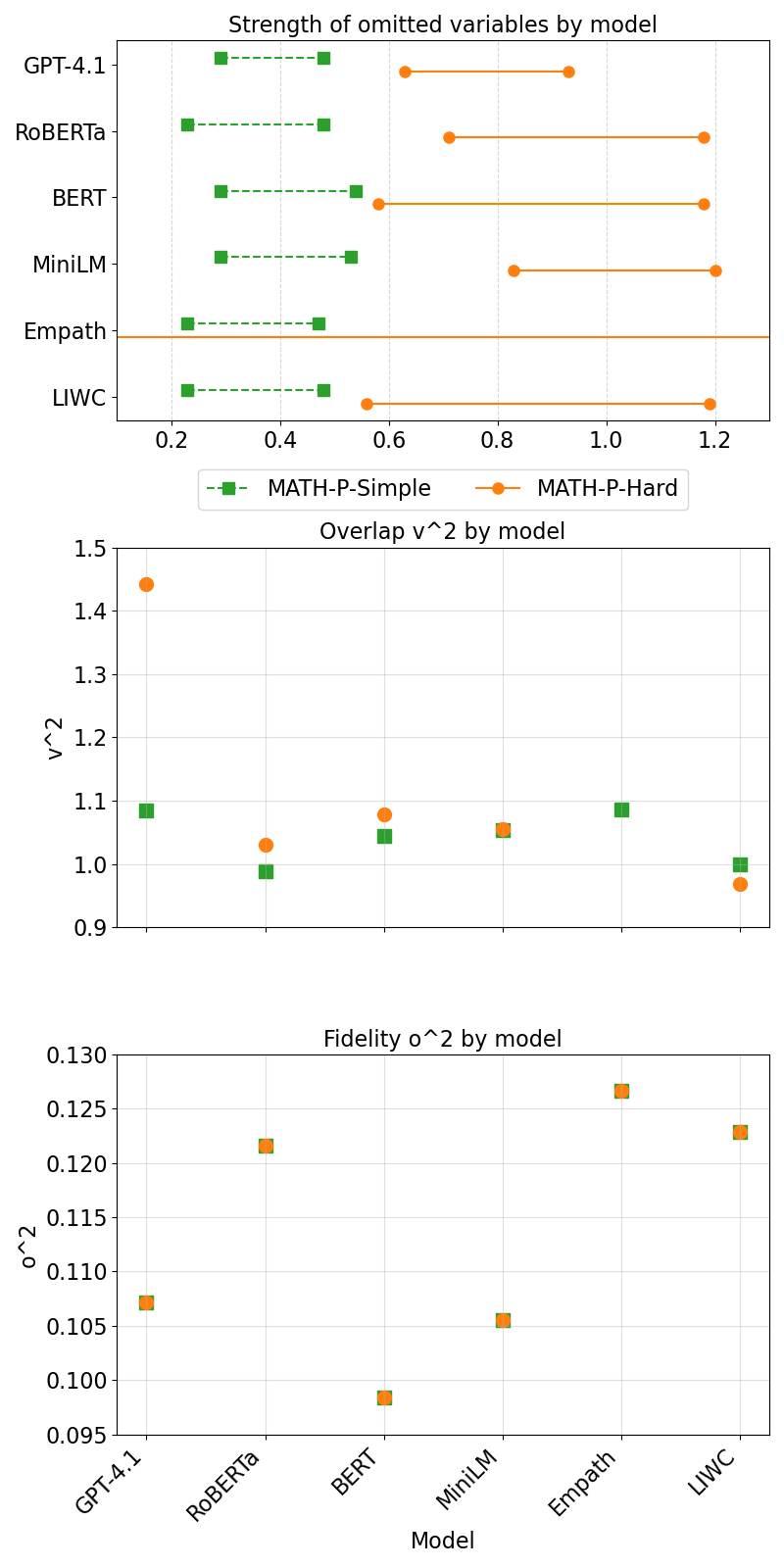}
    \caption{Strength of omitted variables under distribution shift in the Math Reasoning dataset.}
    \label{fig:math}
\end{figure}

Building on this setting, we show that our generalization bound can quantify the strength of omitted variables across language models. This evaluation considers two axes: (i) the true degree of difference between source and target distributions, which sets a ceiling on omitted variable strength, and (ii) the extent to which a model captures those differences, or conversely the strength of the variables it omits.

Across all models, we find that the strength of omitted variables is much larger between MATH and MATH-P-Hard than between MATH and MATH-P-Simple (Figure \ref{fig:math}). Again, this is consistent with expectations, since the differences between MATH and MATH-P-Simple are primarily stylistic, while MATH and MATH-P-Hard differ in their fundamental logic. Language models are likely less capable of capturing this higher-order logic.

Moreover, in MATH-P-Simple, all models exhibit similar sensitivity parameter ranges. Because the differences between MATH and MATH-P-Simple are largely surface-level, even simpler language models seem to be able to capture them from the text, and any remaining omitted-variable effects may instead arise from external factors that do not vary across models. In contrast, when MATH-P-Hard is the target distribution, the models exhibit greater differences in their sensitivity parameter ranges. Unsurprisingly, GPT-4.1---the strongest model---seems least affected by omitted variables, with both the tightest range of sensitivity parameters and the lowest upper bound.
This appears largely due to GPT-4.1's ability to recognize a significant distributional difference between MATH and MATH-P-Hard, which is demonstrated by its much larger $\widehat{\nu}^2$ (Appendix \ref{sec:additional_results}, Figure \ref{fig:math_additional}). The remaining models have $\widehat{\nu}^2$ close to 1,\footnote{Empath in fact has a large negative $\widehat{\nu}^2$, which can indicate a severe violation of the overlap assumption. This violation also results in a large amount of OVB, as evidenced by Empath's very wide sensitivity parameter range.} which indicates that they find almost perfect overlap between MATH and MATH-P-Hard---thereby suggesting that they fail to capture important known differences
between the two distributions.

These measures provide a direct way to assess whether a model has learned a specific concept---such as the logic distinguishing MATH from MATH-P-Hard---without targeted probing tasks or relying on performance metrics, which only indirectly reflect concept knowledge and may be confounded by other model capabilities. In this way, OVB-based comparisons offer interpretable insight into black-box models that would otherwise be opaque.

\subsection{Language Model Optimization}

\begin{figure}[!t]
    \centering
    \includegraphics[width=\columnwidth, 
    ]{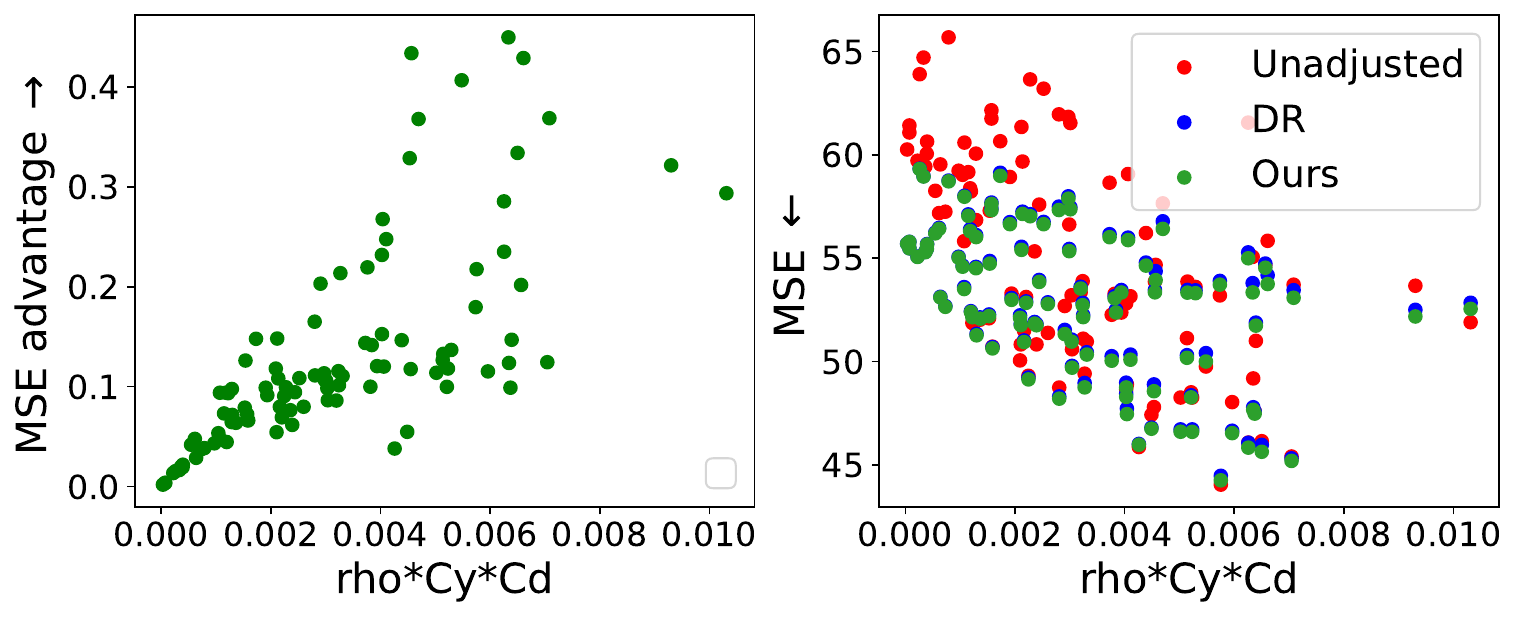}
    \caption{Amazon dataset. On the left, we plot the test performance difference (higher is better) between our approach and a model trained with a standard DR objective against the true strength of omitted variables. On the right, we plot the test performance of our approach and both baselines.}
    \label{fig:amazon}
\end{figure}

\paragraph{Amazon.} Over 100 iterations, we compare the test performance of a model optimized with the worst-case generalization bound against the doubly robust and unadjusted baselines (Figure \ref{fig:amazon}). We recall that in this semi-synthetic setting, we have access to the true sensitivity parameters.

We observe that as the product of the true sensitivity parameters increases---i.e., as the true strength of omitted variables increases---our approach increasingly outperforms the vanilla DR approach with respect to true test performance. This result demonstrates that if sensitivity parameters are appropriately set, (i) our approach is indeed useful for reducing the impact of omitted variables on language model performance, and (ii) the extent to which our approach confers performance advantages over the DR objective increases with the strength of omitted variables.

\begin{figure}[!t]
    \centering
    \begin{subfigure}[t]{0.50\columnwidth}
        \centering
        \includegraphics[
            width=\linewidth,
            trim=0 0 360 0,
            clip
        ]{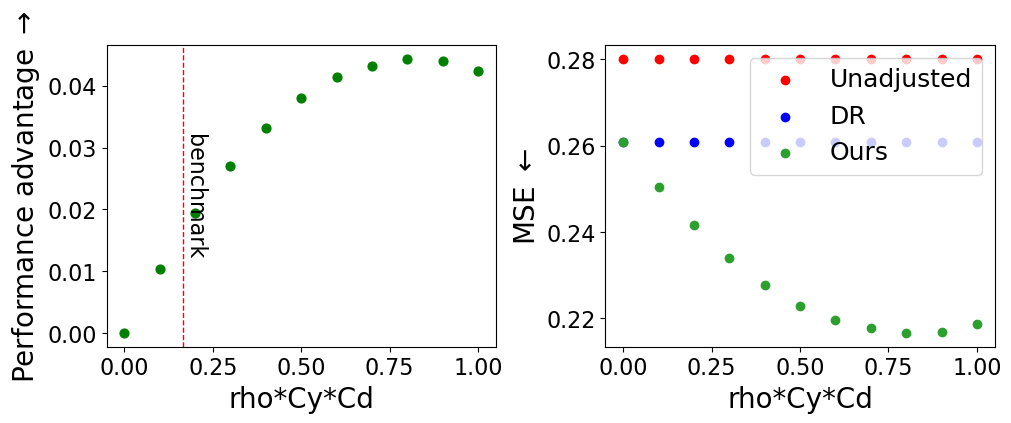}
        \caption{EmoBank (MSE)}
        \label{fig:emobank}
    \end{subfigure}
    \begin{subfigure}[t]{0.47\columnwidth}
        \centering
        \includegraphics[
            width=\linewidth,
            trim=40 0 340 0,
            clip   
        ]{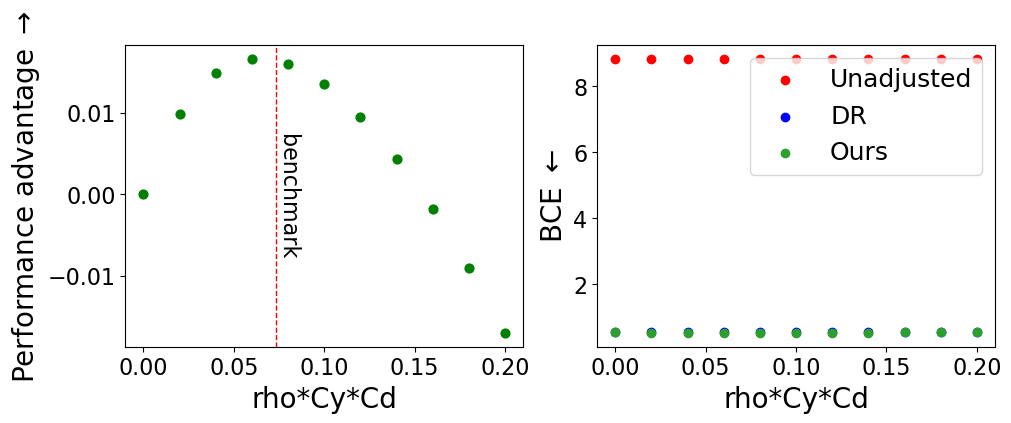}
        \caption{Hate Speech (BCE)}
        \label{fig:hatespeech}
    \end{subfigure}
    \caption{Test performance difference (higher is better) between our approach and a model trained with a standard DR objective against assumed strength of omitted variables.}
    \label{fig:emobank_hatespeech}
\end{figure}

\paragraph{EmoBank and Hate Speech.} In real-world settings where the true sensitivity parameters are unknown, we iterate over 11 plausible values of their product---calibrated using the benchmarked value shown in dotted red---to examine how true test performance varies as the optimization objective becomes more conservative (Figures \ref{fig:emobank} and \ref{fig:hatespeech}). When this product is 0, our method reduces to the DR baseline.

Across both datasets, increasing the sensitivity parameters initially improves true test performance relative to vanilla DR, before performance eventually declines. The gain peaks around $\rho C_YC_D=0.8$ for EmoBank and around $\rho C_YC_D=0.06$ for Hate Speech. These results demonstrate the effectiveness of our approach on real datasets where we have no knowledge of the omitted variables---but also highlight that while accounting for unobserved variables \textit{is} beneficial for model performance under distribution shift, optimizing for an overly pessimistic worst-case generalization bound can be detrimental to model performance.

From these results, we can infer that a relatively large component of distribution shift remains unobserved for EmoBank (that is, the omitted variables are strong), while Hate Speech is much less impacted by unobserved distribution shift. It is possible that most of the distribution shift between Reddit and Gab is already captured in the text representations and therefore observable to language models---or the two distributions may not differ significantly in the first place, which seems plausible since the Reddit portion of the dataset is specifically restricted to subreddits where the user base is expected to be similar to the Gab user base. The distributions of EmoBank, on the other hand, differ over a concrete variable---writer intent---that we know is not directly observable from the data.

\section{Conclusion}

In this paper, we describe a novel unobservable component of distribution shift in language models that can give rise to \textit{omitted variable bias}, compromising model evaluation and optimization. To address this challenge, we present a framework that maps the strength of omitted variables to a bound on the worst-case generalization performance of a language model under distribution shift. We demonstrate three primary contributions that follow from this bound: the ability to more robustly measure models' out-of-distribution performance, to improve true generalization performance relative to standard distribution shift adjustment methods, and to measure the strength of omitted variables when target labels are available.

These contributions point to an interesting line of future work. Although this paper focuses on a distribution shift setting, this approach can be generalized to any language task where the model objective takes or can take a doubly robust form, with the only difference being in how we define the Riesz representer $\alpha$. Therefore, our approach can be used not only to uncover the presence of OVB in other modeling contexts or tasks but can also provide a clear path for mitigating its effects when training language models. More broadly, this suggests a unifying perspective on robustness to omitted variables across a range of language model objectives through their shared doubly robust structure.

\section*{Impact Statement}

This work aims to improve the evaluation and optimization of language models under distribution shift by identifying and accounting for the effects of omitted variable bias. By providing tools to account for uncertainty arising from unobserved variables, our framework may help practitioners make more reliable decisions when deploying models in settings where data distributions are imperfectly observed.

Potential positive impacts include more robust model evaluation, reduced overconfidence in reported performance, and improved reliability of language models in real-world applications. At the same time, as with most advances in model evaluation and optimization, our methods can be applied in domains with varying societal consequences depending on the deployment context. We believe the ethical and societal implications of this work are consistent with those commonly associated with advancing robustness and generalization in machine learning, and we do not foresee specific negative impacts unique to this contribution.

\section*{Acknowledgments}

This material is based upon work partially supported by the \href{https://ror.org/05h1kgg64}{National Institutes of Health} (federal award ID numbers R01MH125740, R01MH132225, and U01MH136535), the \href{https://ror.org/05xpvk416}{National Institute of Standards and Technology} (federal award ID number 6ONANB24D231), and the \href{https://ror.org/05x2bcf33}{Carnegie Mellon University} AI Measurement Science and Engineering Center (AIMSEC). Victoria Lin is partially supported by a Meta Research PhD Fellowship. Any opinions, findings, conclusions, or recommendations expressed in this material are those of the author(s) and do not necessarily reflect the views of the sponsors, and no official endorsement should be inferred.

\bibliography{ref}

\begin{thebibliography}{31}
\providecommand{\natexlab}[1]{#1}
\providecommand{\url}[1]{\texttt{#1}}
\expandafter\ifx\csname urlstyle\endcsname\relax
  \providecommand{\doi}[1]{doi: #1}\else
  \providecommand{\doi}{doi: \begingroup \urlstyle{rm}\Url}\fi

\bibitem[Angrist \& Pischke(2009)Angrist and Pischke]{angrist2009mostly}
Angrist, J.~D. and Pischke, J.-S.
\newblock \emph{Mostly harmless econometrics: An empiricist's companion}.
\newblock Princeton university press, 2009.

\bibitem[Arjovsky et~al.(2020)Arjovsky, Bottou, Gulrajani, and Lopez-Paz]{arjovsky2020invariantriskminimization}
Arjovsky, M., Bottou, L., Gulrajani, I., and Lopez-Paz, D.
\newblock Invariant risk minimization, 2020.
\newblock URL \url{https://arxiv.org/abs/1907.02893}.

\bibitem[Buechel \& Hahn(2017)Buechel and Hahn]{buechel2017emobank}
Buechel, S. and Hahn, U.
\newblock {E}mo{B}ank: Studying the impact of annotation perspective and representation format on dimensional emotion analysis.
\newblock In \emph{Proceedings of the 15th Conference of the {E}uropean Chapter of the Association for Computational Linguistics: Volume 2, Short Papers}, pp.\  578--585, Valencia, Spain, April 2017. Association for Computational Linguistics.
\newblock URL \url{https://aclanthology.org/E17-2092}.

\bibitem[Byrd \& Lipton(2019)Byrd and Lipton]{pmlr-v97-byrd19a}
Byrd, J. and Lipton, Z.
\newblock What is the effect of importance weighting in deep learning?
\newblock In Chaudhuri, K. and Salakhutdinov, R. (eds.), \emph{Proceedings of the 36th International Conference on Machine Learning}, volume~97 of \emph{Proceedings of Machine Learning Research}, pp.\  872--881. PMLR, 09--15 Jun 2019.
\newblock URL \url{https://proceedings.mlr.press/v97/byrd19a.html}.

\bibitem[Chernozhukov et~al.(2024)Chernozhukov, Cinelli, Newey, Sharma, and Syrgkanis]{chernozhukov2024longstoryshortomitted}
Chernozhukov, V., Cinelli, C., Newey, W., Sharma, A., and Syrgkanis, V.
\newblock Long story short: Omitted variable bias in causal machine learning, 2024.
\newblock URL \url{https://arxiv.org/abs/2112.13398}.

\bibitem[Cinelli \& Hazlett(2019)Cinelli and Hazlett]{cinelli2019making}
Cinelli, C. and Hazlett, C.
\newblock {Making Sense of Sensitivity: Extending Omitted Variable Bias}.
\newblock \emph{Journal of the Royal Statistical Society Series B: Statistical Methodology}, 82\penalty0 (1):\penalty0 39--67, 12 2019.
\newblock ISSN 1369-7412.
\newblock \doi{10.1111/rssb.12348}.
\newblock URL \url{https://doi.org/10.1111/rssb.12348}.

\bibitem[Devlin et~al.(2019)Devlin, Chang, Lee, and Toutanova]{devlin-etal-2019-bert}
Devlin, J., Chang, M.-W., Lee, K., and Toutanova, K.
\newblock {BERT}: Pre-training of deep bidirectional transformers for language understanding.
\newblock In Burstein, J., Doran, C., and Solorio, T. (eds.), \emph{Proceedings of the 2019 Conference of the North {A}merican Chapter of the Association for Computational Linguistics: Human Language Technologies, Volume 1 (Long and Short Papers)}, pp.\  4171--4186, Minneapolis, Minnesota, June 2019. Association for Computational Linguistics.
\newblock \doi{10.18653/v1/N19-1423}.
\newblock URL \url{https://aclanthology.org/N19-1423/}.

\bibitem[Duchi \& Namkoong(2021)Duchi and Namkoong]{duchi2021dro}
Duchi, J.~C. and Namkoong, H.
\newblock {Learning models with uniform performance via distributionally robust optimization}.
\newblock \emph{The Annals of Statistics}, 49\penalty0 (3):\penalty0 1378 -- 1406, 2021.
\newblock \doi{10.1214/20-AOS2004}.
\newblock URL \url{https://doi.org/10.1214/20-AOS2004}.

\bibitem[Fast et~al.(2016)Fast, Chen, and Bernstein]{10.1145/2858036.2858535}
Fast, E., Chen, B., and Bernstein, M.~S.
\newblock Empath: Understanding topic signals in large-scale text.
\newblock In \emph{Proceedings of the 2016 CHI Conference on Human Factors in Computing Systems}, CHI '16, pp.\  4647–4657, New York, NY, USA, 2016. Association for Computing Machinery.
\newblock ISBN 9781450333627.
\newblock \doi{10.1145/2858036.2858535}.
\newblock URL \url{https://doi.org/10.1145/2858036.2858535}.

\bibitem[Frank(2000)]{frank2000impact}
Frank, K.~A.
\newblock Impact of a confounding variable on a regression coefficient.
\newblock \emph{Sociological Methods \& Research}, 29\penalty0 (2):\penalty0 147--194, 2000.
\newblock \doi{10.1177/0049124100029002001}.
\newblock URL \url{https://doi.org/10.1177/0049124100029002001}.

\bibitem[Goldberger(1991)]{goldberger1991course}
Goldberger, A.~S.
\newblock \emph{A course in econometrics}.
\newblock Harvard University Press, 1991.

\bibitem[Gonz\'{a}lez \& Nori(2024)Gonz\'{a}lez and Nori]{NEURIPS2024_d5a1f97d}
Gonz\'{a}lez, J. and Nori, A.~V.
\newblock Does reasoning emerge? examining the probabilities of causation in large language models.
\newblock In Globerson, A., Mackey, L., Belgrave, D., Fan, A., Paquet, U., Tomczak, J., and Zhang, C. (eds.), \emph{Advances in Neural Information Processing Systems}, volume~37, pp.\  117737--117761. Curran Associates, Inc., 2024.
\newblock \doi{10.52202/079017-3739}.
\newblock URL \url{https://proceedings.neurips.cc/paper_files/paper/2024/file/d5a1f97d2b922da92e880d13b7d2bf02-Paper-Conference.pdf}.

\bibitem[Hendrycks et~al.(2021)Hendrycks, Burns, Kadavath, Arora, Basart, Tang, Song, and Steinhardt]{hendrycks2021measuring}
Hendrycks, D., Burns, C., Kadavath, S., Arora, A., Basart, S., Tang, E., Song, D., and Steinhardt, J.
\newblock Measuring mathematical problem solving with the {MATH} dataset.
\newblock In \emph{Thirty-fifth Conference on Neural Information Processing Systems Datasets and Benchmarks Track (Round 2)}, 2021.
\newblock URL \url{https://openreview.net/forum?id=7Bywt2mQsCe}.

\bibitem[Huang et~al.(2025)Huang, Guo, Li, Ji, Ge, Li, Guo, Cai, Yuan, Wang, Wu, Yin, Tang, Huang, Jin, Chen, Zhang, and Wang]{huang2025mathperturb}
Huang, K., Guo, J., Li, Z., Ji, X., Ge, J., Li, W., Guo, Y., Cai, T., Yuan, H., Wang, R., Wu, Y., Yin, M., Tang, S., Huang, Y., Jin, C., Chen, X., Zhang, C., and Wang, M.
\newblock {MATH}-perturb: Benchmarking {LLM}s' math reasoning abilities against hard perturbations.
\newblock In \emph{Forty-second International Conference on Machine Learning}, 2025.
\newblock URL \url{https://openreview.net/forum?id=OZy70UggXr}.

\bibitem[H{\"u}y{\"u}k et~al.(2025)H{\"u}y{\"u}k, Xu, Maasch, Nori, and Gonzalez]{huyuk2025reasoning}
H{\"u}y{\"u}k, A., Xu, X., Maasch, J. R. M.~A., Nori, A.~V., and Gonzalez, J.
\newblock Reasoning elicitation in language models via counterfactual feedback.
\newblock In \emph{The Thirteenth International Conference on Learning Representations}, 2025.
\newblock URL \url{https://openreview.net/forum?id=VVixJ9QavY}.

\bibitem[Kallus et~al.(2022)Kallus, Mao, Wang, and Zhou]{pmlr-v162-kallus22a}
Kallus, N., Mao, X., Wang, K., and Zhou, Z.
\newblock Doubly robust distributionally robust off-policy evaluation and learning.
\newblock In Chaudhuri, K., Jegelka, S., Song, L., Szepesvari, C., Niu, G., and Sabato, S. (eds.), \emph{Proceedings of the 39th International Conference on Machine Learning}, volume 162 of \emph{Proceedings of Machine Learning Research}, pp.\  10598--10632. PMLR, 17--23 Jul 2022.
\newblock URL \url{https://proceedings.mlr.press/v162/kallus22a.html}.

\bibitem[Kim et~al.(2022)Kim, Kennedy, and Zubizarreta]{NEURIPS2022_e124f154}
Kim, K., Kennedy, E., and Zubizarreta, J.
\newblock Doubly robust counterfactual classification.
\newblock In Koyejo, S., Mohamed, S., Agarwal, A., Belgrave, D., Cho, K., and Oh, A. (eds.), \emph{Advances in Neural Information Processing Systems}, volume~35, pp.\  34831--34845. Curran Associates, Inc., 2022.
\newblock URL \url{https://proceedings.neurips.cc/paper_files/paper/2022/file/e124f1547f7ac87e33d348b827d4291b-Paper-Conference.pdf}.

\bibitem[Lin \& Morency(2023)Lin and Morency]{lin-morency-2023-sentecon}
Lin, V. and Morency, L.-P.
\newblock {S}ente{C}on: Leveraging lexicons to learn human-interpretable language representations.
\newblock In Rogers, A., Boyd-Graber, J., and Okazaki, N. (eds.), \emph{Findings of the Association for Computational Linguistics: ACL 2023}, pp.\  4312--4331, Toronto, Canada, July 2023. Association for Computational Linguistics.
\newblock \doi{10.18653/v1/2023.findings-acl.264}.
\newblock URL \url{https://aclanthology.org/2023.findings-acl.264}.

\bibitem[Lin et~al.(2024)Lin, Ben-Michael, and Morency]{lin2024optimizing}
Lin, V., Ben-Michael, E., and Morency, L.-P.
\newblock Optimizing language models for human preferences is a causal inference problem.
\newblock In \emph{Proceedings of the Fortieth Conference on Uncertainty in Artificial Intelligence}, UAI '24. JMLR.org, 2024.

\bibitem[Lin et~al.(2025)Lin, Morency, and Ben-Michael]{lin2025isolated}
Lin, V., Morency, L.-P., and Ben-Michael, E.
\newblock Isolated causal effects of natural language.
\newblock In \emph{Forty-second International Conference on Machine Learning}, 2025.
\newblock URL \url{https://openreview.net/forum?id=Z0jnz149L1}.

\bibitem[Liu et~al.(2019)Liu, Ott, Goyal, Du, Joshi, Chen, Levy, Lewis, Zettlemoyer, and Stoyanov]{liu2019robertarobustlyoptimizedbert}
Liu, Y., Ott, M., Goyal, N., Du, J., Joshi, M., Chen, D., Levy, O., Lewis, M., Zettlemoyer, L., and Stoyanov, V.
\newblock Roberta: A robustly optimized bert pretraining approach, 2019.
\newblock URL \url{https://arxiv.org/abs/1907.11692}.

\bibitem[McAuley \& Leskovec(2013)McAuley and Leskovec]{mcauley2013amazon}
McAuley, J. and Leskovec, J.
\newblock Hidden factors and hidden topics: Understanding rating dimensions with review text.
\newblock In \emph{Proceedings of the 7th ACM Conference on Recommender Systems}, RecSys '13, pp.\  165–172, New York, NY, USA, 2013. Association for Computing Machinery.
\newblock ISBN 9781450324090.
\newblock \doi{10.1145/2507157.2507163}.
\newblock URL \url{https://doi.org/10.1145/2507157.2507163}.

\bibitem[OpenAI et~al.(2024)OpenAI, Achiam, Adler, Agarwal, Ahmad, Akkaya, Aleman, Almeida, Altenschmidt, Altman, Anadkat, Avila, Babuschkin, Balaji, Balcom, Baltescu, Bao, Bavarian, Belgum, Bello, Berdine, Bernadett-Shapiro, Berner, Bogdonoff, Boiko, Boyd, Brakman, Brockman, Brooks, Brundage, Button, Cai, Campbell, Cann, Carey, Carlson, Carmichael, Chan, Chang, Chantzis, Chen, Chen, Chen, Chen, Chen, Chess, Cho, Chu, Chung, Cummings, Currier, Dai, Decareaux, Degry, Deutsch, Deville, Dhar, Dohan, Dowling, Dunning, Ecoffet, Eleti, Eloundou, Farhi, Fedus, Felix, Fishman, Forte, Fulford, Gao, Georges, Gibson, Goel, Gogineni, Goh, Gontijo-Lopes, Gordon, Grafstein, Gray, Greene, Gross, Gu, Guo, Hallacy, Han, Harris, He, Heaton, Heidecke, Hesse, Hickey, Hickey, Hoeschele, Houghton, Hsu, Hu, Hu, Huizinga, Jain, Jain, Jang, Jiang, Jiang, Jin, Jin, Jomoto, Jonn, Jun, Kaftan, Łukasz Kaiser, Kamali, Kanitscheider, Keskar, Khan, Kilpatrick, Kim, Kim, Kim, Kirchner, Kiros, Knight, Kokotajlo, Łukasz Kondraciuk, Kondrich,
  Konstantinidis, Kosic, Krueger, Kuo, Lampe, Lan, Lee, Leike, Leung, Levy, Li, Lim, Lin, Lin, Litwin, Lopez, Lowe, Lue, Makanju, Malfacini, Manning, Markov, Markovski, Martin, Mayer, Mayne, McGrew, McKinney, McLeavey, McMillan, McNeil, Medina, Mehta, Menick, Metz, Mishchenko, Mishkin, Monaco, Morikawa, Mossing, Mu, Murati, Murk, Mély, Nair, Nakano, Nayak, Neelakantan, Ngo, Noh, Ouyang, O'Keefe, Pachocki, Paino, Palermo, Pantuliano, Parascandolo, Parish, Parparita, Passos, Pavlov, Peng, Perelman, de~Avila Belbute~Peres, Petrov, de~Oliveira~Pinto, Michael, Pokorny, Pokrass, Pong, Powell, Power, Power, Proehl, Puri, Radford, Rae, Ramesh, Raymond, Real, Rimbach, Ross, Rotsted, Roussez, Ryder, Saltarelli, Sanders, Santurkar, Sastry, Schmidt, Schnurr, Schulman, Selsam, Sheppard, Sherbakov, Shieh, Shoker, Shyam, Sidor, Sigler, Simens, Sitkin, Slama, Sohl, Sokolowsky, Song, Staudacher, Such, Summers, Sutskever, Tang, Tezak, Thompson, Tillet, Tootoonchian, Tseng, Tuggle, Turley, Tworek, Uribe, Vallone, Vijayvergiya,
  Voss, Wainwright, Wang, Wang, Wang, Ward, Wei, Weinmann, Welihinda, Welinder, Weng, Weng, Wiethoff, Willner, Winter, Wolrich, Wong, Workman, Wu, Wu, Wu, Xiao, Xu, Yoo, Yu, Yuan, Zaremba, Zellers, Zhang, Zhang, Zhao, Zheng, Zhuang, Zhuk, and Zoph]{openai2024gpt4technicalreport}
OpenAI, Achiam, J., Adler, S., Agarwal, S., Ahmad, L., Akkaya, I., Aleman, F.~L., Almeida, D., Altenschmidt, J., Altman, S., Anadkat, S., Avila, R., Babuschkin, I., Balaji, S., Balcom, V., Baltescu, P., Bao, H., Bavarian, M., Belgum, J., Bello, I., Berdine, J., Bernadett-Shapiro, G., Berner, C., Bogdonoff, L., Boiko, O., Boyd, M., Brakman, A.-L., Brockman, G., Brooks, T., Brundage, M., Button, K., Cai, T., Campbell, R., Cann, A., Carey, B., Carlson, C., Carmichael, R., Chan, B., Chang, C., Chantzis, F., Chen, D., Chen, S., Chen, R., Chen, J., Chen, M., Chess, B., Cho, C., Chu, C., Chung, H.~W., Cummings, D., Currier, J., Dai, Y., Decareaux, C., Degry, T., Deutsch, N., Deville, D., Dhar, A., Dohan, D., Dowling, S., Dunning, S., Ecoffet, A., Eleti, A., Eloundou, T., Farhi, D., Fedus, L., Felix, N., Fishman, S.~P., Forte, J., Fulford, I., Gao, L., Georges, E., Gibson, C., Goel, V., Gogineni, T., Goh, G., Gontijo-Lopes, R., Gordon, J., Grafstein, M., Gray, S., Greene, R., Gross, J., Gu, S.~S., Guo, Y., Hallacy,
  C., Han, J., Harris, J., He, Y., Heaton, M., Heidecke, J., Hesse, C., Hickey, A., Hickey, W., Hoeschele, P., Houghton, B., Hsu, K., Hu, S., Hu, X., Huizinga, J., Jain, S., Jain, S., Jang, J., Jiang, A., Jiang, R., Jin, H., Jin, D., Jomoto, S., Jonn, B., Jun, H., Kaftan, T., Łukasz Kaiser, Kamali, A., Kanitscheider, I., Keskar, N.~S., Khan, T., Kilpatrick, L., Kim, J.~W., Kim, C., Kim, Y., Kirchner, J.~H., Kiros, J., Knight, M., Kokotajlo, D., Łukasz Kondraciuk, Kondrich, A., Konstantinidis, A., Kosic, K., Krueger, G., Kuo, V., Lampe, M., Lan, I., Lee, T., Leike, J., Leung, J., Levy, D., Li, C.~M., Lim, R., Lin, M., Lin, S., Litwin, M., Lopez, T., Lowe, R., Lue, P., Makanju, A., Malfacini, K., Manning, S., Markov, T., Markovski, Y., Martin, B., Mayer, K., Mayne, A., McGrew, B., McKinney, S.~M., McLeavey, C., McMillan, P., McNeil, J., Medina, D., Mehta, A., Menick, J., Metz, L., Mishchenko, A., Mishkin, P., Monaco, V., Morikawa, E., Mossing, D., Mu, T., Murati, M., Murk, O., Mély, D., Nair, A., Nakano, R.,
  Nayak, R., Neelakantan, A., Ngo, R., Noh, H., Ouyang, L., O'Keefe, C., Pachocki, J., Paino, A., Palermo, J., Pantuliano, A., Parascandolo, G., Parish, J., Parparita, E., Passos, A., Pavlov, M., Peng, A., Perelman, A., de~Avila Belbute~Peres, F., Petrov, M., de~Oliveira~Pinto, H.~P., Michael, Pokorny, Pokrass, M., Pong, V.~H., Powell, T., Power, A., Power, B., Proehl, E., Puri, R., Radford, A., Rae, J., Ramesh, A., Raymond, C., Real, F., Rimbach, K., Ross, C., Rotsted, B., Roussez, H., Ryder, N., Saltarelli, M., Sanders, T., Santurkar, S., Sastry, G., Schmidt, H., Schnurr, D., Schulman, J., Selsam, D., Sheppard, K., Sherbakov, T., Shieh, J., Shoker, S., Shyam, P., Sidor, S., Sigler, E., Simens, M., Sitkin, J., Slama, K., Sohl, I., Sokolowsky, B., Song, Y., Staudacher, N., Such, F.~P., Summers, N., Sutskever, I., Tang, J., Tezak, N., Thompson, M.~B., Tillet, P., Tootoonchian, A., Tseng, E., Tuggle, P., Turley, N., Tworek, J., Uribe, J. F.~C., Vallone, A., Vijayvergiya, A., Voss, C., Wainwright, C., Wang,
  J.~J., Wang, A., Wang, B., Ward, J., Wei, J., Weinmann, C., Welihinda, A., Welinder, P., Weng, J., Weng, L., Wiethoff, M., Willner, D., Winter, C., Wolrich, S., Wong, H., Workman, L., Wu, S., Wu, J., Wu, M., Xiao, K., Xu, T., Yoo, S., Yu, K., Yuan, Q., Zaremba, W., Zellers, R., Zhang, C., Zhang, M., Zhao, S., Zheng, T., Zhuang, J., Zhuk, W., and Zoph, B.
\newblock Gpt-4 technical report, 2024.
\newblock URL \url{https://arxiv.org/abs/2303.08774}.

\bibitem[Oster(2019)]{oster2019unobservable}
Oster, E.
\newblock Unobservable selection and coefficient stability: Theory and evidence.
\newblock \emph{Journal of Business \& Economic Statistics}, 37\penalty0 (2):\penalty0 187--204, 2019.
\newblock \doi{10.1080/07350015.2016.1227711}.
\newblock URL \url{https://doi.org/10.1080/07350015.2016.1227711}.

\bibitem[Pennebaker et~al.(2015)Pennebaker, Boyd, Jordan, and Blackburn]{pennebaker2015development}
Pennebaker, J.~W., Boyd, R.~L., Jordan, K., and Blackburn, K.
\newblock The development and psychometric properties of liwc2015.
\newblock Technical report, 2015.
\newblock URL \url{http://liwc.net/LIWC2007LanguageManual.pdf}.

\bibitem[Qian et~al.(2019)Qian, Bethke, Liu, Belding, and Wang]{qian-etal-2019-benchmark}
Qian, J., Bethke, A., Liu, Y., Belding, E., and Wang, W.~Y.
\newblock A benchmark dataset for learning to intervene in online hate speech.
\newblock In \emph{Proceedings of the 2019 Conference on Empirical Methods in Natural Language Processing and the 9th International Joint Conference on Natural Language Processing (EMNLP-IJCNLP)}, pp.\  4755--4764, Hong Kong, China, November 2019. Association for Computational Linguistics.
\newblock \doi{10.18653/v1/D19-1482}.
\newblock URL \url{https://aclanthology.org/D19-1482}.

\bibitem[Robins et~al.(1994)Robins, Rotnitzky, and Ping~Zhao]{Robins1994}
Robins, J.~M., Rotnitzky, A., and Ping~Zhao, L.
\newblock Estimation of {Regression} {Coefficients} {When} {Some} {Regressors} are not {Always} {Observed}.
\newblock \emph{Journal of the American Statistical Association}, 89\penalty0 (427):\penalty0 846--866, 1994.

\bibitem[Song et~al.(2020)Song, Tan, Qin, Lu, and Liu]{NEURIPS2020_c3a690be}
Song, K., Tan, X., Qin, T., Lu, J., and Liu, T.-Y.
\newblock Mpnet: Masked and permuted pre-training for language understanding.
\newblock In Larochelle, H., Ranzato, M., Hadsell, R., Balcan, M., and Lin, H. (eds.), \emph{Advances in Neural Information Processing Systems}, volume~33, pp.\  16857--16867. Curran Associates, Inc., 2020.
\newblock URL \url{https://proceedings.neurips.cc/paper_files/paper/2020/file/c3a690be93aa602ee2dc0ccab5b7b67e-Paper.pdf}.

\bibitem[Wang et~al.(2020)Wang, Wei, Dong, Bao, Yang, and Zhou]{NEURIPS2020_3f5ee243}
Wang, W., Wei, F., Dong, L., Bao, H., Yang, N., and Zhou, M.
\newblock Minilm: Deep self-attention distillation for task-agnostic compression of pre-trained transformers.
\newblock In Larochelle, H., Ranzato, M., Hadsell, R., Balcan, M., and Lin, H. (eds.), \emph{Advances in Neural Information Processing Systems}, volume~33, pp.\  5776--5788. Curran Associates, Inc., 2020.
\newblock URL \url{https://proceedings.neurips.cc/paper_files/paper/2020/file/3f5ee243547dee91fbd053c1c4a845aa-Paper.pdf}.

\bibitem[Xu et~al.(2025)Xu, Lawrence, Dubey, Pandey, Ueno, Falck, Nori, Sharma, Sharma, and Gonzalez]{xu2025reimagine}
Xu, X., Lawrence, R., Dubey, K., Pandey, A., Ueno, R., Falck, F., Nori, A.~V., Sharma, R., Sharma, A., and Gonzalez, J.
\newblock {RE}-{IMAGINE}: Symbolic benchmark synthesis for reasoning evaluation.
\newblock In \emph{Forty-second International Conference on Machine Learning}, 2025.
\newblock URL \url{https://openreview.net/forum?id=QJPl0DWajD}.

\bibitem[Yang et~al.(2023)Yang, Zhang, Katabi, and Ghassemi]{10.5555/3618408.3620060}
Yang, Y., Zhang, H., Katabi, D., and Ghassemi, M.
\newblock Change is hard: a closer look at subpopulation shift.
\newblock In \emph{Proceedings of the 40th International Conference on Machine Learning}, ICML'23. JMLR.org, 2023.

\end{thebibliography}
\bibliographystyle{icml2026}

\newpage
\appendix
\onecolumn
\section{Derivations}
\subsection{General Doubly Robust Objective}
\label{sec:dr_derivation}

Recall the covariate shift assumption $\mathbb E_Q[\ell(Y, X;f) \mid X, Z]=\mathbb E_Q[\ell(Y, X;f) \mid X, Z]$ and that $g(X, Z;f)=\mathbb E[\ell(Y, X;f) \mid X, Z]$.
\begin{align*}
\mathcal L
    &= \mathbb E_Q[\ell(Y,X;f)] \\
    &= \mathbb E_Q\left[
    \mathbb E_Q[\ell(Y,X;f)\mid X,Z]
    \right] \\
    &= \mathbb E_Q\left[
    \mathbb E_P[\ell(Y,X;f)\mid X,Z]
    \right] \\
    &= \mathbb E_Q\left[g(X, Z;f)\right]
\end{align*}
And

\begin{align*}
\mathcal L
    &= \mathbb E_Q[\ell(Y,X;f)] \\
    &= \mathbb E_P\left[\frac{dQ}{dP}(X,Z) \ell(Y, X;f)\right] \\
    &= \mathcal{L}_\text{IPW}
\end{align*}
So
\begin{align*}
\mathcal L &= \mathbb E_Q\left[g(X, Z;f)\right] + \mathbb E_P\left[\frac{dQ}{dP}(X,Z) (g(X, Z; f) - \ell(Y, X;f))\right] \\
&= \mathcal L_\text{DR}
\end{align*}

\subsection{GLM Doubly Robust Objective}
\label{sec:glm_derivations}
Recall the covariate shift assumption $\mathbb E_Q[Y \mid X, Z]=\mathbb E_Q[Y \mid X, Z]$ and that $g(X, Z;f)=\mathbb E[Y \mid X, Z]$.
\[
\mathcal{L} = \mathbb E_Q[-Y\cdot\eta(X;f)+b(\eta(X;f))].
\]

Since $ \mathbb E_Q[b(\eta(X;f))]$ does not depend on the labels $Y$, we can focus on $\mathbb E_Q[-Y\cdot\eta(X;f)$.

\begin{align*}
    \mathbb E_Q[-Y\cdot\eta(X;f)] = \mathbb E_P\left[-\frac{dQ}{dP}(X,Z) Y\cdot \eta(X;f) \right],
\end{align*}
and 
\begin{align*}
 \mathbb E_Q[-Y\cdot\eta(X;f)]  & =  \mathbb E_Q[- \mathbb E_Q[Y \mid X, Z] \cdot\eta(X;f)] \\
 & = \mathbb E_Q[-g(X,Z) \cdot\eta(X;f)]\\
\end{align*}
So

\[
\mathcal{L}_\text{DR} =  \mathbb E_Q[b(\eta(X;f)) - \eta(X;f)g(X, Z)] - \mathbb E_P[\alpha(X, Z) \eta(X;f)(Y-g(X, Z))]
\]

\section{GLM Log-Likelihood Losses}
\label{sec:glm_losses}

\subsection{Regression}

Recall that we have $\eta(x;f)=f(x)$, where $f(x) \in \mathbb{R}$ is the scalar prediction from the model, and $b(\eta(x;f))=\frac{1}{2}\eta(x;f)^2$. Then letting $Y \in \mathbb{R}$ denote the label, the doubly robust loss for regression is:
\[
    \mathcal{L}_{\text{DR}}=\mathbb{E}_Q[f(X)^2 - 2f(X)g(X, Z)] -2\mathbb{E}_P\left[f(X)\frac{\text{Pr}_Q(X, Z)}{\text{Pr}_P(X, Z)}(Y-g(X, Z))\right]
\]

\subsection{Binary Classification}

Recall that we have $\eta(x;f)=f(x)$, where $f(x) \in \mathbb{R}$ is the logit for the positive class, and $b(\eta(x;f))=\log(1+\exp(\eta(x;f)))$. Then letting $Y \in \{0, 1\}$ denote the label, the doubly robust loss for binary classification is:
\begin{equation*}
\begin{split}
\mathcal{L}_{\text{DR}}&=-\mathbb{E}_P\left[\frac{\text{Pr}_Q(X, Z)}{\text{Pr}_P(X, Z)}\log\left(\frac{f(X)}{1-f(X)}\right)(Y-g(X,Z))\right] - \mathbb{E}_Q[g(X, Z)\log f(X)+(1-g(X, Z))\log (1-f(X))]
\end{split}
\end{equation*}

\subsection{Multiclass Classification}

Recall that we have $\eta(x;f)=f(x)$, where $f(x) \in \mathbb{R}^K$ is the logit vector for the $K$ classes, and $b(\eta(x;f))=\log\sum_{j=1}^K \exp(\eta_j(x;f))$. Then letting $Y \in \{0, 1\}^K$ denote the one-hot label vector and $g_k(X,Z)=\mathbb E[Y_k \mid X, Z]$, the doubly robust loss for binary classification is:
\[
    \mathcal{L}_{\text{DR}}=-\mathbb{E}_P\left[\sum_{k=1}^V \frac{\text{Pr}_Q(X, Z)}{\text{Pr}_P(X, Z)}\log f_k(X)(Y_k-g_k(X, Z))\right] - \mathbb{E}_Q\left[\sum_{k=1}^V \log f_k(X)g_k(X, Z)\right]
\]

\subsection{Text Generation}
Recall that we have $\eta(x;f)=f^{(t)}(x)$, where $f^{(t)}(x) \in \mathbb{R}^K$ is the logit vector for the $K$ tokens in the vocabulary, and $b(\eta(x;f))=\sum_{t=1}^T\log\sum_{j=1}^K \exp(\eta_j^{(t)}(x;f))$. Then letting $Y^{(t)} \in \{0, 1\}^K$ denote the one-hot label vector for the next token at time $t$ and $g_k^{(t)}(X,Z)=\mathbb E[Y_k^{(t)} \mid X, Z]$, the doubly robust loss for binary classification is:
\[
    \mathcal{L}_{\text{DR}}=-\mathbb{E}_P\left[\sum_{t=1}^T\sum_{k=1}^K \frac{\text{Pr}_Q(X, Z)}{\text{Pr}_P(X, Z)}\log f_k^{(t)}(X)(Y_k^{(t)}-g_k^{(t)}(X, Z))\right] - \mathbb{E}_Q\left[\sum_{t=1}^T\sum_{k=1}^K \log f_k^{(t)}(X)g_k^{(t)}(X, Z)\right]
\]

\section{Experiments}
\label{sec:appendix_experiments}

\subsection{Language Model Implementation}
\label{sec:appendix_language-reps}

\begin{table}[!h]
    \centering
    \caption{Technical details for language model implementations.}
    \vskip 0.1in
    \begin{tabular}{c|c|c|c}
    \toprule
         & Language & Library & Model \\
    \midrule
        LIWC & Python & \verb|liwc| & - \\
        Empath & Python & \verb|empath| & -\\
        SenteCon & Python & \verb|sentecon| & - \\
        BERT & Python & \makecell{\Verb|transformers|} & \Verb|bert-base-uncased| \\
        RoBERTa & Python & \makecell{\Verb|transformers|} & \Verb|roberta-base| \\
        MiniLM & Python & \makecell{\Verb|sentence-transformers|} & \Verb|all-MiniLM-L6-v2| \\
        MPNet & Python & \makecell{\Verb|sentence-transformers|} & \Verb|all-mpnet-base-v2|\\
        GPT-4.1 & - & - & \Verb|gpt-4.1| \\
    \bottomrule
    \end{tabular}
    \label{tab:language_reps}
\end{table}

To implement our lexicons, we use the third-party \verb|liwc| Python library and the \verb|empath| library provided by its creators. SenteCon-LIWC and SenteCon-Empath representations are generated using the \verb|sentecon| library. BERT and RoBERTa embeddings are obtained via the HuggingFace \Verb|transformers| library using the pre-trained models \Verb|bert-base-uncased| and \Verb|roberta-base|, respectively, while MPNet and MiniLM embeddings are obtained via the HuggingFace \verb|sentence-transformers| library with the pre-trained models \verb|all-mpnet-base-v2| and \verb|all-MiniLM-L6-v2|. Finally, results from GPT-4.1 were retrieved through OpenAI's Batch API.

Additional technical details are provided in Table \ref{tab:language_reps}.

\subsection{Computing Resources}
\label{sec:computing}
All experiments were conducted on consumer-level machines using consumer-level NVIDIA GPUs.

\newpage
\section{Additional Experimental Results}
\label{sec:additional_results}

\begin{figure}[!ht]
    \centering
    \includegraphics[width=0.5\columnwidth, trim=0 0 0 380, clip]{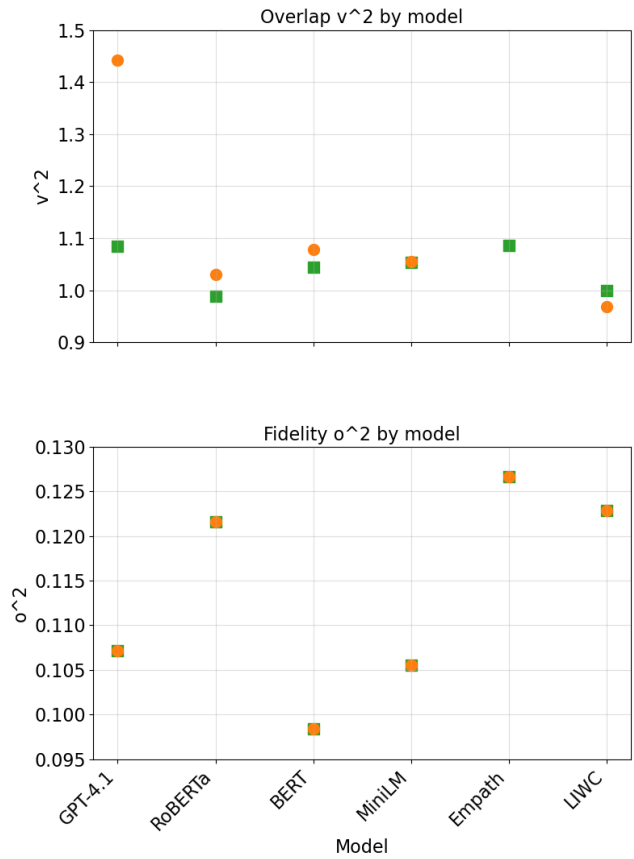}
    \caption{OVB metrics under distribution shift in the Math Reasoning dataset.}
    \label{fig:math_additional}
\end{figure}

\begin{figure}[!ht]
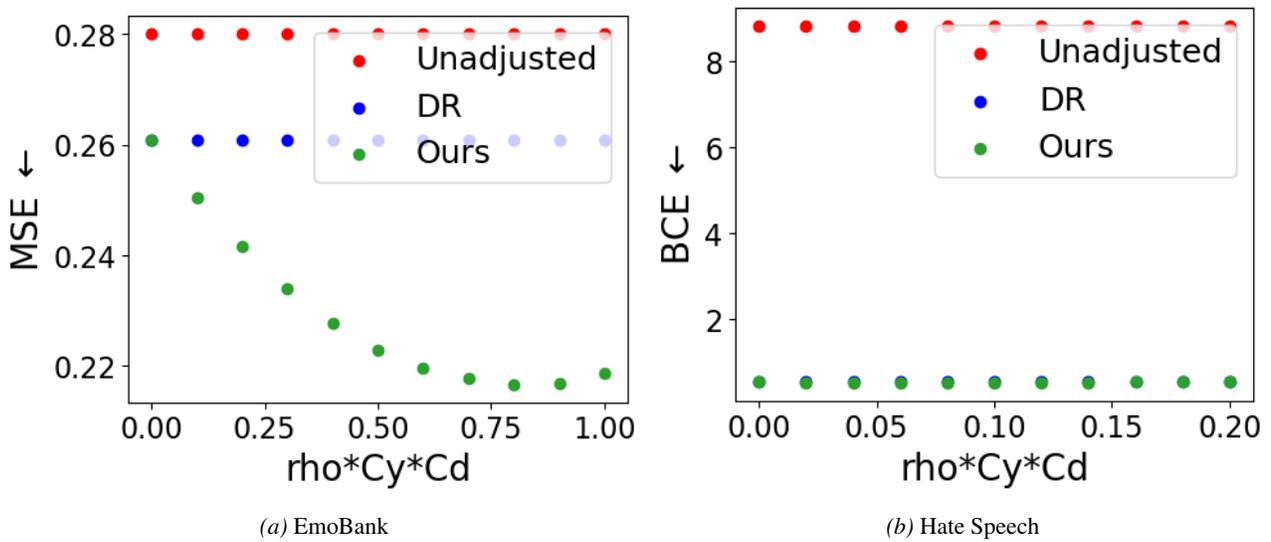

    \centering
    \begin{subfigure}[t]{0.50\columnwidth}
        \centering
        \includegraphics[
            width=\linewidth,
            trim=365 0 0 0,
            clip
        ]{figures/numerical_mse_liwc_Cy.png}
        \caption{EmoBank}
        \label{fig:emobank}
    \end{subfigure}
    \begin{subfigure}[t]{0.48\columnwidth}
        \centering
        \includegraphics[
            width=\linewidth,
            trim=385 0 0 0,
            clip   
        ]{figures/binary_bce_sentecon_Cd.png}
        \caption{Hate Speech}
        \label{fig:hatespeech}
    \end{subfigure}
    \caption{Test performances of our approach, the DR baseline, and the unadjusted baseline plotted against strength of omitted variables.}
    \label{fig:emobank_hatespeech}
\end{figure}

\end{document}